\documentclass[journal]{IEEEtran}

\usepackage[left=0.625in,right=0.625in,top=0.75in,bottom=1in]{geometry}

\usepackage{amsmath}
\usepackage{amssymb}

\usepackage{graphicx}
\usepackage{array}
\usepackage{multirow}
\usepackage{tabularx}
\usepackage{caption}
\usepackage{subcaption}
\usepackage{float}
\usepackage{dblfloatfix}

\usepackage{algorithm}
\usepackage{algpseudocode}
\usepackage{xcolor}
\definecolor{mypink2}{RGB}{0,0,255}
\definecolor{green}{RGB}{0,128,0}
\newcommand{\red}[1]{\textcolor{red}{#1}}

\newcommand{\vlm}{Qwen2-VL-7B-Instruct}  
\newcommand{\vlmshort}{Qwen2-VL-7B}       
\newcommand{\vlmbackend}{mlx-vlm, locally hosted} 
\newcommand{\vlmres}{1024$\times$768}   

\newcommand{\nframes}{41} 
\usepackage{cite}
\usepackage{url}
\usepackage{hyperref}
\hypersetup{hidelinks}

\usepackage{ragged2e}
\usepackage{verbatim}
\usepackage{lipsum}
\usepackage{multicol}
\usepackage{blindtext}
\usepackage[utf8]{inputenc}
\usepackage[acronym]{glossaries}
\usepackage[most]{tcolorbox}

\begin{document}

\title{\fontsize{16pt}{16pt}\selectfont LAPF: LLM-Agent-Based Path Finder Using the UAVScenes Dataset}

\author{ Yousef~Emami,~\IEEEmembership{Senior Member,~IEEE,}
 Mohammadhossein~Homaei,~\IEEEmembership{Senior Member,~IEEE,}
        Hao~Zhou,~\IEEEmembership{Senior Member,~IEEE,}
Miguel~Gutiérrez~Gaitán,~\IEEEmembership{Senior Member,~IEEE,} 
 and~Atefeh~Hajijamali Arani,~\IEEEmembership{Member,~IEEE,
      }
and~Rui~Zhang,~\IEEEmembership{Fellow,~IEEE,
      }

\thanks{Copyright (c) 2026 IEEE. Personal use of this material is permitted. However, permission to use this material for any other purposes must be obtained from the IEEE by sending a request to pubs-permissions@ieee.org.}    
        
}       
\maketitle

\begin{abstract}

 Unmanned aerial vehicles (UAVs) are increasingly deployed for autonomous navigation in complex outdoor environments, where dynamic conditions and mission requirements require intelligent adaptive decision-making. Existing optimization-based, Machine Learning (ML), and Reinforcement Learning (RL) approaches often rely on predefined models or task-specific training, limiting their generalization and adaptability in uncertain scenarios. Recent Large Language Model (LLM)-assisted approaches offer promising reasoning capabilities but remain constrained by limited agentic functionality, including insufficient memory, planning, and tool interaction mechanisms. This paper proposes an LLM-Agent-Based Path Finder (LAPF) framework for autonomous UAV navigation in town-scale outdoor environments. LAPF extends LLM-assisted navigation by integrating perception, memory, planning, and action modules into a closed-loop cognitive architecture. The proposed agent leverages prior navigation experiences, performs Chain-of-Thought (CoT) reasoning, couples each detected hazard to a bounded corrective action, and dynamically refines waypoint decisions based on environmental feedback. The three independent trials per method demonstrate that LAPF achieves mean path lengths of $512.83$\,m and $506.37$\,m compared to the straight-line optimum $497.33$\,m corresponding to path length reductions of $17.2\%$ and $15.6\%$ relative to CoT prompting and absolute path efficiencies of $97.1\%$ and $98.1\%$ in open-field and obstacle-injected scenarios, respectively. Furthermore, LAPF is the only evaluated approach that couples every detected hazard to a bounded, metric-neutral corrective action while maintaining near-goal stability with zero clamp events in both scenarios, whereas CoT prompting increases from $9.7$ to $14.0$ events. 
\end{abstract}

\IEEEpeerreviewmaketitle

\section{Introduction}

Uncrewed Aerial Vehicles (UAVs) provide three-dimensional mobility, dynamic Line-of-Sight (LoS) establishment, and high data throughput, making them suitable for operation in complex outdoor environments such as public safety, environmental monitoring, agriculture, and parcel delivery~\cite{10246260,11458872}. Their ability to dynamically adjust trajectories according to mission requirements enables flexible coverage in dense urban areas where terrestrial infrastructure may be obstructed or unavailable. Furthermore, UAVs can serve as aerial base stations, data collectors, relay platforms, and sensing nodes, extending their role from simple transportation platforms to intelligent networked systems\cite{11241785}.
\begin{figure} \centering \includegraphics[width=1\linewidth]{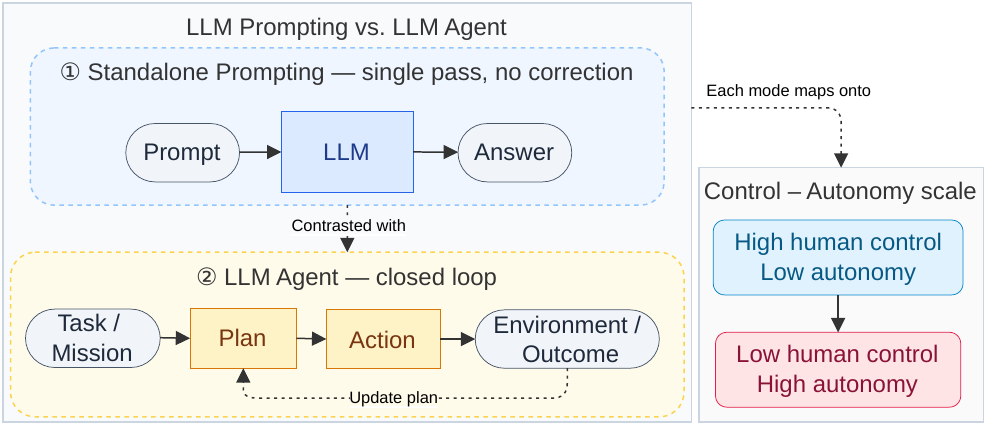} \caption{ Architectural comparison between standalone LLM prompting and LLM agents. Conventional prompting (top) executes a single prompt-to-answer pass without autonomous correction capabilities. In contrast, the LLM agent (bottom) establishes a closed-loop system via an internal Plan$\rightarrow$Action cycle and an \emph{Update plan} feedback mechanism. As illustrated on the right, these paradigms map directly onto a control-autonomy scale, highlighting the trade-off between direct human control and system autonomy. } \label{fig:llm_vs_agent} \end{figure}
UAV navigation and communication systems have traditionally relied on optimization-based methods and, more recently, Machine Learning (ML) and Reinforcement Learning (RL) approaches to address challenges such as trajectory planning, resource allocation, and autonomous decision-making \cite{10612249}. Although these methods have achieved significant progress, their effectiveness is often limited by predefined system models, task-specific training requirements, and insufficient generalization to dynamic and uncertain environments. In particular, emergency response, cooperative UAV networking, and large-scale autonomous operations require decision-making mechanisms that can rapidly adapt to changing conditions without extensive retraining or manual redesign.
\par
Large Language Models (LLMs) have recently demonstrated remarkable capabilities in reasoning, contextual understanding, and task automation across domains such as finance, healthcare, and networking~\cite{10822885,raiaan2024review}. Techniques like In-Context Learning (ICL) and Chain-of-Thought (CoT) reasoning~\cite{wei2022chain} enable LLMs to solve complex tasks by leveraging contextual information and structured reasoning processes. These emerging capabilities have motivated their exploration in UAV and autonomous system applications~\cite{emami2026llm,el2025lightweight}. However, directly applying standalone LLMs to UAV navigation remains challenging due to three fundamental limitations:

\begin{enumerate}
    \item \textbf{Multimodal perception gap}: Conventional text-based LLMs cannot directly interpret heterogeneous UAV sensory inputs, including RGB images, LiDAR point clouds, depth information, and 6-Degrees of Freedom (DoF) localization data, which are critical for reliable navigation. Addressing this gap requires a backbone capable of ingesting visual evidence natively, together with an interface that renders the non-visual channels in a form the backbone can reason over \cite {10495592}.
    
    \item \textbf{Limited agentic capabilities}: Autonomous UAV operation requires memory, adaptive planning, long-term spatial reasoning, and continual interaction with the environment, which are beyond the capabilities of conventional single-pass LLM prompting\cite{11480623}.
    
    \item \textbf{Lack of tool integration}: Practical navigation relies on external modules, including simulators, obstacle detection systems, localization tools, and trajectory planners. Standalone LLM prompting lacks mechanisms for reliable tool invocation and closed-loop execution\cite{ferrag2026llm}.
\end{enumerate}

Moreover, existing UAV navigation approaches based on optimization, ML, and RL often require extensive environment modeling or task-specific training, offer limited interpretability, and struggle to integrate heterogeneous perception information into a unified decision-making framework~\cite{Jingwen2026,11540994}. These limitations motivate the development of more flexible cognitive architectures capable of combining perception, reasoning, memory, and action. LLM agents offer a promising solution by extending conventional LLMs with additional cognitive modules, including memory, planning, tool use, and action execution. As illustrated in Fig.~\ref{fig:llm_vs_agent}, unlike standalone prompting approaches that perform a single prompt-to-response operation, LLM agents establish a closed-loop interaction process consisting of perception, reasoning, planning, execution, and feedback-based refinement. Such architectures enable autonomous systems to adapt their decisions based on environmental changes, retrieve prior experiences, and interact with external computational resources. Therefore, LLM agents provide a potential pathway toward intelligent UAV navigation with enhanced situational awareness, cross-layer coordination, and long-horizon decision-making capabilities. Nevertheless, several challenges remain, including reliable reasoning, real-time operation, domain-specific adaptation, and integration with existing UAV control frameworks\cite{zhou2024large,ferrag2026llm,jiang2026large}.

Motivated by these opportunities, this work proposes the LLM-Agent-Based Path Finder (LAPF), a cognitive navigation framework for UAVs designed to enable adaptive waypoint generation and autonomous replanning. LAPF is evaluated using the UAVScenes dataset~\cite{wang20256}, whose town-scale outdoor sequences provide a realistic evaluation environment for multimodal UAV perception and navigation. UAVScenes is a multimodal UAV perception benchmark combining RGB images, LiDAR, 6-DoF poses, semantic labels, and 3D maps.
It supports diverse tasks such as segmentation, localization, place recognition, depth estimation, and novel view synthesis. The dataset enables evaluation of advanced learning-based and LLM-assisted UAV navigation systems. The LAPF agent integrates perception, memory, planning, and action modules to perform CoT reasoning, retrieve previous navigation experiences, invoke external tools such as obstacle checkers and trajectory validators, and generate candidate waypoints. During execution, newly observed environmental changes can trigger autonomous replanning, enabling continuous refinement of navigation decisions through a closed-loop perception–reasoning–action process.

\par

\par
The main contributions of this paper are summarized as follows:
\begin{itemize}
 \item LAPF, an LLM-agent-based autonomous UAV navigation framework, extends conventional LLM prompting by integrating multimodal perception, episodic memory, and tool-enforced action within a closed-loop architecture. It grounds reasoning in UAVScenes observations by combining native RGB input to an open-weight Vision-Language Model (VLM) with symbolic LiDAR and 6-DoF pose representations in a single inference call. Running entirely on a consumer laptop, LAPF demonstrates that its performance gains stem from the agent architecture rather than a proprietary backbone. The framework leverages memory and external tools for adaptive waypoint generation, safety validation, and trajectory refinement. Its two-tier Action module guarantees safety-critical hazard checks at every decision step while allowing discretionary tool use, ensuring hazards always trigger corrective actions, a reliability capability absent in prompting-only methods. Consequently, LAPF enables adaptive perception–reasoning–action cycles without requiring additional model fine-tuning.
 \item The effectiveness of LAPF is validated on the UAVScenes AMtown03 sequence under open-field and obstacle-injected scenarios. Using the same \vlmshort{}~\cite{wang2024qwen2vl}  backbone and identical multimodal observation sequences for all methods, LAPF reduces mean path length by $17.2\%$ and $15.6\%$ relative to CoT prompting across the two scenarios, reaches path efficiencies of $97.05\%$ and $98.09\%$ respectively, and maintains zero near-goal clamp events in both, indicating that its advantage is preserved rather than eroded as environmental complexity increases. 
    
\end{itemize}

The remainder of this paper is organized as follows. Section~\ref{sec3} reviews related work. Section~\ref{sec4} describes the system model, and Section~\ref{sec:problem} formulates the path-planning objective. Section~\ref{sec5} details the proposed LAPF framework. Section~\ref{sec6} presents the numerical results and discussion. Finally, Section~\ref{sec7} concludes the paper.

\begin{table*}[t]
\centering
\caption{Comparative Analysis of LAPF Against Related UAV and LLM-Agent Frameworks}
\label{tab:lapf_comparison}
\renewcommand{\arraystretch}{1.3}

\resizebox{\textwidth}{!}{%
\begin{tabular}{p{2.5cm}p{6cm}cccccc}
\hline
\textbf{Framework} & \textbf{Domain} &
\textbf{\begin{tabular}[c]{@{}c@{}}Multimodal\\ Input\end{tabular}} &
\textbf{\begin{tabular}[c]{@{}c@{}}Explicit\\ Memory\end{tabular}} &
\textbf{\begin{tabular}[c]{@{}c@{}}Tool\\ Use\end{tabular}} &
\textbf{\begin{tabular}[c]{@{}c@{}}Fine-tuning\\ Required\end{tabular}} &
\textbf{\begin{tabular}[c]{@{}c@{}}UAV-\\specific\end{tabular}} &
\textbf{\begin{tabular}[c]{@{}c@{}}Real Sensor\\ Fusion\end{tabular}} \\
\hline

\cite{10012880} & Urban UAV safety path planning & $\checkmark$ & $\times$ & $\times$ & $\times$ & $\checkmark$ & $\times$ \\

\cite{11470760} & Urban UAV obstacle avoidance & $\times$ & $\times$ & $\times$ & $\times$ & $\checkmark$ & $\times$ \\

\cite{9196558} & Urban UAV reconstruction & $\checkmark$ & $\times$ & $\times$ & $\times$ & $\checkmark$ & $\checkmark$ \\

\cite{9165709} & Risk-aware UAV path planning & $\checkmark$ & $\times$ & $\times$ & $\times$ & $\checkmark$ & $\times$ \\

\cite{drones8120701} & UAV optimization-based path planning (LGPSO) & $\times$ & $\times$ & $\times$ & $\times$ & $\checkmark$ & $\times$ \\

\cite{zhang2019new} & GNSS-aware urban UAV navigation & $\checkmark$ & $\times$ & $\times$ & $\times$ & $\checkmark$ & $\checkmark$ \\

\cite{Jingwen2026} & Wireless slicing & $\times$ & $\checkmark$ & $\checkmark$ & $\times$ & $\times$ & $\times$ \\

\cite{Yu2026} & Wireless multi-task optimization & $\times$ & $\times$ & $\times$ & $\checkmark$ & $\times$ & $\times$ \\

\cite{Chen20266G} & 6G network engineering & $\times$ & $\checkmark$ & $\checkmark$ & $\checkmark$ & $\times$ & $\times$ \\

\cite{Tong2026w} & Wireless agent benchmark & $\times$ & $\times$ & $\checkmark$ & -- & $\times$ & $\times$ \\

\cite{emami2026llm} & LLM-UAV general framework & -- & -- & -- & -- & $\checkmark$ & -- \\

\cite{11184156} & UAV scheduling & $\times$ & $\times$ & $\times$ & $\times$ & $\checkmark$ & $\times$ \\

\cite{11433611} & UAV trajectory and velocity optimization & $\times$ & $\times$ & $\times$ & $\times$ & $\checkmark$ & $\times$ \\

\cite{11399950} & UAV AoI and flight resource allocation & $\times$ & $\times$ & $\times$ & $\times$ & $\checkmark$ & $\times$ \\
\cite{sun2026autofly} & UAV autonomous navigation (AutoFly) 
& $\checkmark$ & $\times$ & $\times$ & $\checkmark$ & $\checkmark$ & $\checkmark$ \\

\cite{zhou2026freefly} & UAV outdoor VLN (FreeFly-thinking) 
& $\checkmark$ & $\times$ & $\times$ & $\checkmark$ & $\checkmark$ & $\times$ \\

\cite{dong2025se} & Self-evolving VLN (SE-VLN) 
& $\checkmark$ & $\checkmark$ & $\times$ & $\checkmark$ & $\times$ & $\times$ \\
\textbf{LAPF (this work)} & UAV navigation & $\checkmark$ & $\checkmark$ & $\checkmark$ & $\times$ & $\checkmark$ & $\checkmark^{\dagger}$ \\

\hline
\end{tabular}%
}

\vspace{2pt}
\raggedright
\footnotesize $^{\dagger}$Real recorded UAVScenes streams. The RGB frame is consumed natively by the VLM backbone; the LiDAR return and 6-DoF pose enter as a symbolic state at the perception interface.

\footnotesize ``--'' denotes not reported or not applicable.
\end{table*}

\section{Related Work} \label{sec3}
This section reviews classical UAV path-planning approaches, recent LLM-agent frameworks for wireless networks, and emerging LLM-assisted UAV navigation methods. Finally, we identify the limitations of existing approaches and highlight the research gap that motivates the proposed LAPF framework.

\subsection{Classical Path Planning}
Iwashina \emph{et al.}\cite{10012880}, Sui \emph{et al.}\cite{11470760}, and Kuang \emph{et al.}\cite{9196558} study UAV path planning in urban environments, focusing on safety, efficiency, and autonomy. Iwashina \emph{et al.} propose a V2X-based dynamic risk map with RRT for collision-aware navigation in changing urban conditions. Sui et al. apply RRT to obstacle-dense environments and demonstrate faster computation, greater adaptability, and smoother paths compared with A* and ACO. Kuang et al. develop a framework for real-time urban mapping that combines path generation, height estimation, SLAM-based image acquisition, and coverage prediction to improve reconstruction quality and reduce operational time.

Hu \emph{et al.}\cite{9165709} suggest a comprehensive UAV risk assessment framework for safe urban operations by quantifying collision risks involving people, vehicles, and manned aircraft. Their approach integrates risk costs into UAV path planning by generating risk maps and optimizing flight trajectories. Cheng \emph{et al.}\cite{drones8120701} present an improved particle swarm optimization-based UAV path planning algorithm (LGPSO) to address limitations of traditional PSO, such as convergence to local optima and instability in complex urban environments. By incorporating safety constraints, stochastic inertia weights, asymmetric learning factors, Lévy flight strategies, and a greedy evaluation mechanism, the proposed method enhances path optimization and search efficiency. Zhang \emph{et al.}\cite{zhang2019new} address the challenge of Global Navigation Satellite System (GNSS) positioning errors in urban UAV operations caused by multipath effects and Non-Line-of-Sight (NLOS) conditions. They propose a risk-aware path planning approach that predicts GNSS errors using ray-tracing simulations and 3D building models, then integrates the predicted positioning error into an enhanced A* path planning algorithm. The results show that accounting for localization uncertainty during route planning enables UAVs to select safer paths with fewer positioning errors compared with conventional distance-based planning methods.

\subsection{LLM Agent for Wireless Networks}
Tong \emph{et al.} \cite{Jingwen2026} introduce WirelessAgent, an LLM-agent framework for intelligent wireless network management built on four modules: perception, memory, planning, and action. Using LangGraph-based agent workflows, it enables autonomous decision-making and achieves higher bandwidth utilization than prompt-based methods while approaching the performance of rule-based optimization. Yu \emph{et al.} \cite{Yu2026} present WirelessLLM-Agent, a multi-task wireless optimization framework that uses semantic state serialization and Mixture-of-Experts (MoE)-LoRA for efficient knowledge sharing. A two-stage training strategy that combines supervised fine-tuning with Group Relative Policy Optimization (GRPO)-based RL improves long-term decision-making. Chen \emph{et al.}\cite{Chen20266G} develop 6GAgentGym and 6G-Forge, a closed-loop tool-based environment for training and evaluating wireless agents through NS-3 simulations, enabling small open-source models to achieve competitive performance on long-horizon 6G tasks. To address the limitations of existing benchmarks, Tong \emph{et al.} \cite{Tong2026w} introduce a tolerance-aware benchmark covering reasoning, resource allocation, and mobility-aware decisions, with tool-based evaluation and catastrophic-error detection to identify practical agent failures.

\subsection{LLMs for UAV Networks}
Emami \emph{et al.} have investigated the integration of LLMs with UAV networks through a series of contributions that establish a foundation for intelligent UAV decision-making. In their survey \cite{emami2026llm}, they present a unified framework for LLM-assisted UAV systems, summarizing existing architectures, methodologies, and applications, and highlighting the potential of LLMs as high-level reasoning modules for autonomous aerial networks. Building on this perspective, they propose ICL-based solutions for UAV missions, including Search-and-Rescue (SAR) operations, where LLM-assisted scheduling enables adaptive data collection by jointly considering dynamic channel conditions, battery constraints, and queue states \cite{11184156}. Extending the role of LLMs from task scheduling to real-time control, they further investigate public safety applications by employing LLM-assisted ICL for UAV trajectory planning and adaptive velocity control \cite{11433611}. To address freshness-aware sensing scenarios, they introduce FRSICL, an online resource allocation framework that integrates LLM-assisted reasoning with optimization techniques to jointly determine data collection schedules and UAV velocities, thereby minimizing the average Age of Information for ground sensors \cite{11399950}.

Beyond pure LLM-assisted reasoning, recent works have explored the integration of multimodal foundation models to enable UAVs with richer perception and embodied intelligence. Sun \emph{et al.}\cite{sun2026autofly} introduce AutoFly, an end-to-end Vision-Language-Action (VLA) model that enables UAVs to navigate unknown environments using coarse guidance rather than predefined trajectories. By incorporating pseudo-depth spatial reasoning, progressive multimodal training, and a real-world autonomous navigation dataset, AutoFly improves UAV decision-making and achieves a 3.9\% higher success rate than existing VLA baselines. Similarly, Zhou \emph{et al.}\cite{zhou2026freefly} suggest an end-to-end Vision-Language Navigation framework that enables UAVs to perform outdoor navigation through explicit reasoning over visual observations and language instructions. Their framework combines a dedicated UAV navigation dataset with CoT reasoning and a two-stage training strategy consisting of supervised fine-tuning and reinforcement learning, improving robustness and generalization in unseen environments. Going beyond static model deployment, Dong \emph{et al.}\cite{dong2025se} introduce a self-evolving Vision-Language Navigation framework that allows agents to continuously improve during testing through experience-driven learning. By integrating hierarchical memory, retrieval-augmented reasoning, and reflection mechanisms, their approach enhances long-term adaptation and achieves significant performance improvements in unseen environments. 
\par
In contrast, among the reviewed approaches, LAPF combines explicit memory, tool-enforced action, and no fine-tuning requirement. It is also grounded in multimodal UAV data (RGB, LiDAR, and 6-DoF pose) from a real-world UAV dataset (UAVScenes), rather than synthetic states. The reviewed wireless-domain agents show the value of memory and tool use. However, they lack multimodal grounding and aerial safety-critical design. UAV-specific LLM works avoid fine-tuning through prompting alone, but lack both memory of prior outcomes and a tool-use path to act on an unsafe classification. LAPF's Action module addresses this gap. Classical planners remain efficient for narrowly defined geometric objectives, such as risk-aware navigation, obstacle avoidance, and coverage mapping. They do not natively interpret natural-language mission intent, retain memory across missions, or jointly reason over heterogeneous constraints without planner redesign. LAPF is designed to complement rather than replace classical planners.
\par
Overall, the reviewed literature reveals a consistent gap: existing LLM-agent frameworks are either agentic but domain-general (equipped with memory and/or tool use, but validated on wireless network management rather than physical navigation, and without multimodal sensor grounding), or UAV-specific but architecturally shallow (effective at single-shot or in-context optimization of scheduling and trajectory parameters, but reliant on prompting alone, with no persistent memory of past navigation outcomes and no enforced mechanism to translate a detected hazard into corrective action). To the best of our knowledge, few UAV navigation systems combine multimodal grounding, episodic memory, and tool-enforced safety control in a unified LLM-agent architecture. LAPF is designed to address this gap by combining these capabilities within a closed-loop UAV navigation architecture. Table \ref{tab:lapf_comparison} provides a comparative analysis.

\begin{figure}[t]
    \centering
    \captionsetup{justification=justified}
    \includegraphics[width=1\linewidth]{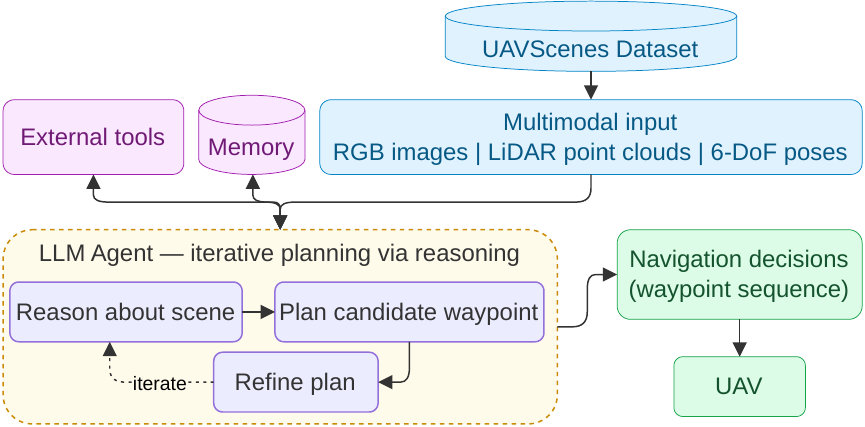}
    \caption{A typical UAV path planning pipeline with LLM agents. The agent receives multimodal input from the UAVScenes dataset and, through an iterative Reason$\rightarrow$Plan$\rightarrow$Refine reasoning cycle, leverages external tools and memory in a bidirectional exchange to generate the UAV's navigation decisions, one waypoint per decision step.}
\label{fig:typical_uav_llm_agent}
\end{figure}

\section{System Model} \label{sec4}

This section presents the system model of the proposed LAPF framework for UAV navigation. The system consists of a single UAV equipped with a multi‑modal sensor suite, an onboard or edge‑based LLM agent, with the UAVScenes dataset serving as the source of environmental observations for evaluation, as depicted in Fig. \ref{fig:typical_uav_llm_agent}. The proposed LAPF enables adaptive path planning through closed‑loop interaction between the UAV and the LLM agent.

The UAV is equipped with a synchronized camera–LiDAR configuration, matching that used in UAVScenes and comprising a Livox‑Avia LiDAR and a high‑resolution RGB camera (\(2448\times2048\) pixels at 10 Hz) as used in UAVScenes. This combination provides dense 2D texture information and accurate 3D spatial measurements. The system also includes GNSS/inertial measurement unit (IMU) measurements with Real-time Kinematic (RTK) support, which contributes to 6‑DoF pose estimation (3‑DoF position and 3‑DoF orientation). All sensor streams are hardware‑synchronised to produce frame‑wise aligned image–LiDAR pairs.

The UAV operates in environments represented in the UAVScenes dataset, which include towns, valleys, airports, and islands. However, extension to dense urban environments remains future work because UAVScenes currently
does not include high‑rise buildings, narrow streets, and pedestrian crowds. The UAV is equipped with:

\begin{itemize}
    \item An RGB camera capturing high‑resolution images ($2448 \times 2048$ pixels) at 10 Hz, synchronized with the LiDAR.
    \item A Livox Avia LiDAR sensor providing frame‑wise 3D point clouds at 10 Hz.
    \item A hardware‑synchronized camera‑LiDAR suite ensuring pixel‑wise alignment via pre‑calibrated extrinsics.
    \item An onboard localization system that supplies 6-DoF poses, reconstructed using structure from motion (SfM) in DJI Terra and refined from the original 4‑DoF GNSS/RTK data.
\end{itemize}
These sensors generate multi‑modal observations at each time step $t$, denoted as:

\begin{equation}
\mathcal{O}_t^{\mathrm{full}} = \{ \mathcal{I}_t, \mathcal{P}_t, \mathcal{D}_t, \mathcal{S}_t, \mathcal{L}_t \}
\label{eq:sensor_suite}
\end{equation}

where
\begin{itemize}
    \item $\mathcal{I}_t$ is the RGB image.
    \item $\mathcal{P}_t$ is the raw LiDAR point cloud (Livox Avia, after noise filtering).
    \item $\mathcal{D}_t$ is the dense depth map derived from projecting $\mathcal{P}_t$ onto the image plane using the known calibration.
    \item $\mathcal{S}_t$ is the semantic segmentation mask, comprising 16 static classes (e.g., road, building, and vegetation), 2 dynamic object classes (sedan and truck only; pedestrians and other small dynamic objects are not annotated), and 1 background class – for a total of 19 classes. Annotations are provided for both the image and the LiDAR point cloud on a per‑frame basis.
        \item $\mathcal{L}_t$ is the 6‑DoF pose $(x, y, z, \psi, \theta, \phi)$ (position and Euler angles) reconstructed via SfM and aligned with the 3D map. The median positional accuracy on benchmark splits is approximately 6.1 m for state‑of‑the‑art methods, indicating that the benchmarked methods do not achieve submeter localization accuracy.
\end{itemize}

Eq.~\eqref{eq:sensor_suite} specifies the full sensor suite available in UAVScenes; the subset instantiated in this work is denoted by $\mathcal{O}_t$ and defined in \eqref{eq:obs_model}, as summarized in Table~\ref{tab:instantiation}. The three instantiated channels reach the backbone through two distinct pathways. The RGB frame $\mathcal{I}_t$ is registered, time-matched and resampled to \vlmres{}, and is then passed \emph{natively} to the VLM backbone as an image input, so that visual scene evidence is available to the reasoning stage without an intermediate captioning step. The LiDAR point cloud and the pose, by contrast, are not visual quantities and are rendered symbolically: $\mathcal{P}_t$ enters through the summary statistic $\sigma_t$ of \eqref{eq:lidar_summary} rather than as raw points, and the pose enters as $\mathcal{L}_t$. Fusion of the visual and symbolic pathways therefore occurs inside the backbone at inference time, whereas the LiDAR return is reduced to a geometric abstraction at the interface. This separation is deliberate: it keeps the safety-critical clearance test of \eqref{eq:hazard_indicator} on a deterministic geometric channel, independent of the backbone's visual inference, while still exposing the scene appearance to the planner. The dense depth map $\mathcal{D}_t$ and semantic mask $\mathcal{S}_t$, although annotated in UAVScenes, are not supplied in this work.

\begin{table}[t]
\centering
\caption{Instantiation of the system model in the present implementation}
\label{tab:instantiation}
\renewcommand{\arraystretch}{1.15}
\begin{tabular}{p{1.05cm}p{2.5cm}p{3.6cm}}
\hline
\textbf{Symbol} & \textbf{System model} & \textbf{Instantiation} \\ \hline
$\mathcal{I}_t$ & RGB image & Passed natively to the VLM backbone at \vlmres{}; registered and time-matched \\
$\mathcal{P}_t$ & Raw point cloud & Entered via the summary $\sigma_t$,
\eqref{eq:lidar_summary} \\
$\mathcal{L}_t$ & 6-DoF pose & Pose $(R_t,p_t)$ with the time-matched RTK fix \\
$\mathcal{D}_t$ & Dense depth map & Not fused (future work) \\
$\mathcal{S}_t$ & Semantic mask & Not fused (future work) \\
$M_t$ & Episodic memory & $k{=}3$ most recent episodes, recency-based \\
$h_t$ & Hazard indicator & $\mathbb{1}[c_t\le d_{\mathrm{safe}}]$,
\eqref{eq:hazard_indicator} \\
$\mathcal{R}$ & Corrective operator & Deterministic contraction toward
$p_{\mathrm{goal}}$, \eqref{eq:replan} \\ \hline
\end{tabular}
\end{table}

\begin{figure*}
    \centering
    \includegraphics[width=0.85\linewidth]{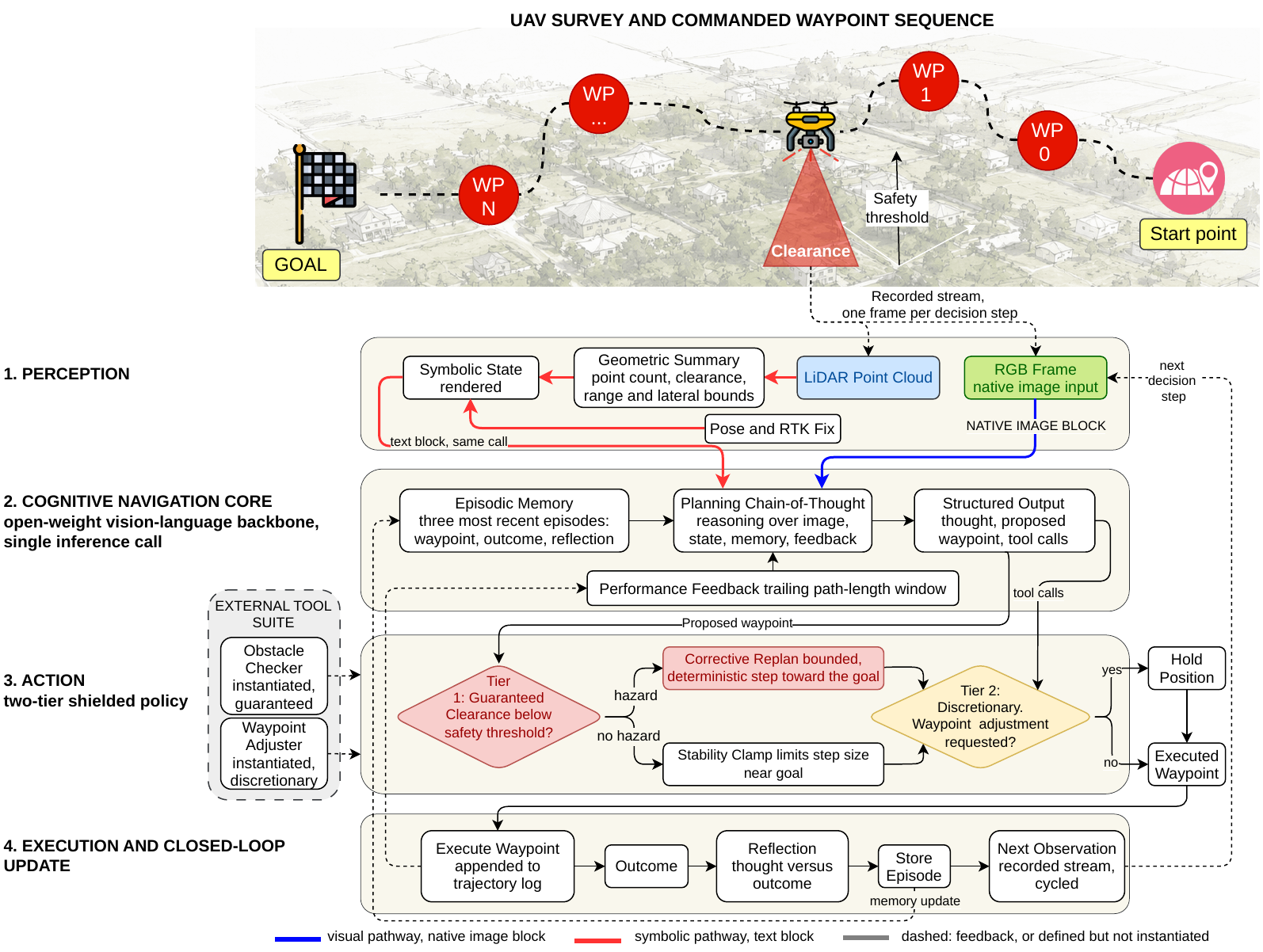}
    \caption{Overview of the proposed LAPF framework. The observation reaches the backbone through two distinct pathways: the RGB frame is passed natively as an image block and is
never textualized, whereas the LiDAR return and the pose are reduced to a symbolic state; fusion of the two pathways occurs inside the backbone at inference time, within a single
multimodal call. The two-tier Action module evaluates the clearance test with guaranteed invocation at every decision step, so that a registered violation is always coupled to a bounded
corrective replan, which is metric-neutral by construction; the discretionary tier remains under agent authority.}
    \label{fig:digital567}
\end{figure*}

\section{Problem Formulation} \label{sec:problem}
This section formulates the UAV navigation problem addressed by LAPF. We first define the observation model and sequential decision process, followed by the memory-conditioned decision policy and the enforced hazard-response mechanism. Finally, we present the optimization objective and the performance metrics used to evaluate the proposed framework.
\par
This paper aims to minimize the projected mission length $\widetilde{L}$ of \eqref{eq:path_length} while ensuring an enforced hazard response during autonomous navigation in the considered environment. The proposed LAPF leverages multimodal environmental information obtained from the UAVScenes dataset, including RGB images, summarized LiDAR returns, and 6-DoF localization states, to enable path planning through an LLM-assisted autonomous agent. The UAV agent dynamically reasons about obstacle distributions, traversability conditions, localization confidence, and mission objectives to generate adaptive navigation trajectories.

\subsection{Decision Process and Observation Model}
The mission is modelled as a discrete-time, partially observable sequential
decision process over a horizon $T_{\max}$. At decision step $t$ the agent
holds a commanded state $\mathrm{wp}_t\in\mathbb{R}^3$, initialized at
$\mathrm{wp}_0=p_0$, and receives an observation
\begin{equation}
\mathcal{O}_t=\bigl\{\mathcal{I}_t,\;\sigma_t,\;\mathcal{L}_t\bigr\},
\label{eq:obs_model}
\end{equation}
where $\mathcal{I}_t$ is the RGB frame, $\mathcal{L}_t=(R_t,p_t)\in SE(3)$ is
the 6-DoF pose, and $\sigma_t$ is a compact statistic of the LiDAR return,
\begin{equation}
\!\!\sigma_t=\bigl(n_t,\,r_t^{\min},\,r_t^{\max},\,b_t^{\min},\,b_t^{\max},\, \!c_t\bigr),
\qquad
c_t=\min_{i}\;x_i ,
\label{eq:lidar_summary}
\end{equation}

in which $x_i$ is the boresight coordinate of return $\mathbf{p}_i$, $n_t$ is the return count, $[r_t^{\min},r_t^{\max}]$ are the bounds of the Euclidean range, $[b_t^{\min},b_t^{\max}]$ are the lateral bounds in the plane orthogonal to the boresight, and $c_t$ is the smallest boresight coordinate. Since $c_t\le\lVert \mathbf{p}_i\rVert$ for the corresponding return $\mathbf{p}_i$, $c_t$ lower-bounds the Euclidean range. The test in \eqref{eq:hazard_indicator} is therefore conservative by construction and cannot under-declare a hazard.

The Livox-Avia unit in UAVScenes is nadir-mounted and hardware-aligned with the RGB camera, so $c_t$ is the clearance between the platform and the nearest surface below it. Across the evaluated sequence the returns span a $77^{\circ}$ cone, matching the nominal sensor field of view, and the boresight coordinate is tightly concentrated within each frame, as expected for a downward survey over near-planar terrain. The sequence is flown at survey altitude throughout, and the sensed clearance $c_t$ remains above $15$\,m in every frame, the lowest values arising from tree canopy.

\emph{Scope of the safety test.} The enforcement mechanism of
Section~\ref{sec:problem}-C is defined over a scalar hazard signal and is
agnostic to the channel supplying it. Whatever geometric quantity is placed
in $\sigma_t$, its registration is guaranteed to be coupled to the bounded
response of \eqref{eq:replan}. This is a property of the action layer rather
than of the sensor, and it is what prompting-only pipelines lack.

The channel instantiated here is the boresight clearance $c_t$, which
measures separation from the nearest surface below the platform. Two
properties of this channel are stated explicitly. First, $c_t$ certifies
clearance along the boresight and not along the direction of commanded
motion, so \eqref{eq:enforcement_condition} guarantees a prescribed response
to a registered clearance violation rather than horizontal collision
avoidance. This is consistent with the disclaimer accompanying Property~1.
Second, the return resolves range but not bearing within the sensed
footprint, so a waypoint-level clearance $\lVert \mathrm{wp}-o\rVert$ cannot
be evaluated without assuming a bearing that $\sigma_t$ does not certify.
The test is consequently expressed at the level of the observation, which is
the conservative choice under this abstraction.

This scope follows from the UAVScenes sensor configuration rather than from
LAPF. Under the common-observation protocol it applies identically to all
three evaluated methods and therefore does not affect the comparison.
Lateral scene structure remains available to the planner through the nadir
RGB frame, which the VLM backbone consumes directly. That channel is
excluded from enforcement because a safety-critical guarantee cannot depend
on a stochastic visual inference. Extending enforcement to a
bearing-resolved channel is identified as future work.

\subsection{Memory-Conditioned Decision Policy}
Let $M_t=\bigl\{(\mathcal{O}_j,\mathrm{wp}_j,y_j,\rho_j)\bigr\}_{j=t-k}^{t-1}$ denote the episodic memory of the $k$ most recent decision episodes, where $y_j$ is the execution outcome and $\rho_j$ the generated reflection, and let $\phi_t$ be the scalar performance-feedback signal derived from the trailing window of path-length statistics. The agent implements the policy \begin{equation} \pi:\;\bigl(\mathcal{P}_{sys},\,\mathcal{G},\,\mathcal{O}_t,\,M_t,\,\phi_t\bigr) \;\longmapsto\; \bigl(\widehat{\mathrm{wp}}_t,\;\mathrm{Tools}_t\bigr), \label{eq:policy} \end{equation} where $\mathcal{G}$ is the mission specification, $\widehat{\mathrm{wp}}_t$ the proposed waypoint and $\mathrm{Tools}_t$ the requested tool set.

\subsection{Enforced Hazard Response and Stability Regulation}
The proposed waypoint $\widehat{\mathrm{wp}}_t$ is not executed directly. Given the
minimum admissible clearance $d_{\mathrm{safe}}$, the hazard indicator
\begin{equation}
h_t=\mathbb{1}\bigl[\,c_t\le d_{\mathrm{safe}}\,\bigr]
\label{eq:hazard_indicator}
\end{equation}
is evaluated at every decision step with guaranteed invocation, so that its
activation does not depend on whether the corresponding tool is named in a
particular sampling of $\mathrm{Tools}_t$;here, $\mathbb{1}[\cdot]$ denotes the indicator function, the agent retains full authority
over $\widehat{\mathrm{wp}}_t$ and over the discretionary tier of
$\mathrm{Tools}_t$. When $h_t=1$, the model-authored proposal is discarded and replaced by the deterministic corrective operator
\begin{equation}
\mathcal{R}(\mathrm{wp}_{t-1})=\mathrm{wp}_{t-1}
+\kappa\bigl(p_{\mathrm{goal}}-\mathrm{wp}_{t-1}\bigr),
\qquad \kappa\in(0,1),
\label{eq:replan}
\end{equation}
which is a function of the commanded state and the mission goal alone. This operator serves strictly as a low-level fail-safe guardrail rather than an intelligent planner. Its role is not to optimize the trajectory, but to provide a prescribed and bounded reaction in critical states. The cognitive value of the framework lies in the agent's ability to use memory and continuous reasoning to anticipate hazards, thereby learning to navigate efficiently to minimize the activation of this deterministic shield. When triggered, the executed displacement becomes $\lVert\mathcal{R}-\mathrm{wp}_{t-1}\rVert=
\kappa\lVert p_{\mathrm{goal}}-\mathrm{wp}_{t-1}\rVert$, which is bounded and reproducible across trials, and independent of temperature sampling, in
contrast to an arbitrary model-authored coordinate. The magnitude $\kappa\lVert p_{\mathrm{goal}}-\mathrm{wp}_{t-1}\rVert$ is a commanded set-point rather than an instantaneous displacement: the operator specifies where the vehicle is directed once a hazard is registered, and the rate at which that set-point is tracked is delegated to the low-level flight controller, which is outside the scope of the decision layer modelled here. Enforcing a decision-layer displacement bound jointly with $\mathcal{C}$, so that set-point and kinematic feasibility are regulated by a single mechanism, is identified as future work. We state explicitly that
$\mathcal{R}$ enforces a bounded and deterministic response to a detected
hazard, and not a geometric clearance guarantee, which
\eqref{eq:lidar_summary} does not support. Although $\mathcal{R}$
contracts toward $p_{\mathrm{goal}}$, it is not an implicit optimizer of the
reported objective: Property~1 below shows that it leaves $\widetilde{L}_t$
exactly invariant.

A second operator regulates step magnitude in the terminal phase, where
independently resampled coordinates would otherwise be charged as travelled
distance. For a goal tolerance $\varepsilon$ and a maximum near-goal
displacement $\delta_{\max}$,
\begin{equation}
\mathcal{C}(w)=
\begin{cases}
\mathrm{wp}_{t-1}+\delta_{\max}
\dfrac{w-\mathrm{wp}_{t-1}}
{\lVert w-\mathrm{wp}_{t-1}\rVert},
&
\substack{
\lVert \mathrm{wp}_{t-1}-p_{\mathrm{goal}}\rVert<\varepsilon,\\
\lVert w-\mathrm{wp}_{t-1}\rVert>\delta_{\max}
},
\\[6pt]
w, & \text{otherwise}.
\end{cases}
\label{eq:clamp}
\end{equation}
together with a hold-position operator, invoked when the agent selects
\texttt{WaypointAdjust},
\begin{equation}
\mathcal{H}(w)=
\begin{cases}
\mathrm{wp}_{t-1}, &
\lVert \mathrm{wp}_{t-1}-p_{\mathrm{goal}}\rVert<\varepsilon,\\
w, & \text{otherwise.}
\end{cases}
\label{eq:hold}
\end{equation}
The executed waypoint is obtained by the ordered composition
\begin{equation}
\begin{aligned}
\mathrm{wp}_t &=
\mathcal{H}^{\,\mathbb{1}[\texttt{WaypointAdjust}\in\mathrm{Tools}_t]}
\Bigl(
h_t\,\mathcal{R}(\mathrm{wp}_{t-1}) \\
&\qquad +(1-h_t)\,\mathcal{C}\bigl(\widehat{\mathrm{wp}}_t\bigr)
\Bigr).
\end{aligned}
\label{eq:executed_wp}
\end{equation}
Equation~\eqref{eq:executed_wp} defines a \emph{shielded} policy
$\pi_{\mathrm{safe}}$: the language model retains authority over the waypoint
on every step classified as hazard-free, while $\mathcal{C}$ constrains only
model-authored proposals and is deliberately not applied to $\mathcal{R}$,
giving hazard response strict precedence over step-magnitude regulation.

\subsection{Objective and Reported Estimators}
Let $L_t=\sum_{i=1}^{t}\lVert \mathrm{wp}_i-\mathrm{wp}_{i-1}\rVert$ denote the
travelled length and
$L^{\star}=\lVert p_{\mathrm{goal}}-\mathrm{wp}_0\rVert$ the straight-line
optimum. The projected mission length, path efficiency, detour ratio and clamp
count are defined as
\begin{equation}
\widetilde{L}_t=L_t+\lVert p_{\mathrm{goal}}-\mathrm{wp}_t\rVert,
\qquad
\eta_t=\frac{L^{\star}}{\widetilde{L}_t},
\qquad
\varrho_t=\eta_t^{-1},
\label{eq:path_length}
\end{equation}
\begin{equation}
N_c=\sum_{t=1}^{T_{\max}}
\mathbb{1}\bigl[\,\mathcal{C}\ \text{is active at step } t\,\bigr],
\qquad
N_h=\sum_{t=1}^{T_{\max}} h_t .
\label{eq:clamp_count}
\end{equation}
The estimator $\widetilde{L}_t$ augments the travelled length with the residual
distance to the goal, so that a trajectory is not credited for terminating
early away from $p_{\mathrm{goal}}$. By the triangle inequality
$\widetilde{L}_t\ge L^{\star}$, hence $\eta_t\le 1$, with equality attained
exactly when the commanded sequence is collinear with, and monotone along, the
start--goal segment. Reported efficiencies are means of the per-trial ratios $\eta_t$ rather than the ratio of the mean path lengths; by Jensen's inequality the two differ slightly.

\emph{Property 1 (Metric neutrality of $\mathcal{R}$).} Let
$\Delta_t=\widetilde{L}_t-\widetilde{L}_{t-1}$. The triangle inequality gives
$\Delta_t\ge0$, with equality if and only if $\mathrm{wp}_t$ lies on the
segment $[\mathrm{wp}_{t-1},p_{\mathrm{goal}}]$, which \eqref{eq:replan} places
it on for every $\kappa\in(0,1)$. Since $\widetilde{L}_0=L^{\star}$, this
yields the exact decomposition
\begin{equation}
\widetilde{L}_{T_{\max}}=L^{\star}+\!\!\sum_{t\,:\,h_t=0}\!\!\Delta_t ,
\qquad \Delta_t\ge0 .
\label{eq:excess_decomposition}
\end{equation}
Hence the enforcement pathway contributes identically zero to $\eta_t$
and is scored through $N_h$ rather than through $\eta_t$. This is a designed
separation of concerns: the shield is required to be metric-neutral so that
safety enforcement and trajectory quality are never conflated. The decomposition
\eqref{eq:excess_decomposition} is an exact identity on the realized
trajectory and so invokes no counterfactual. Neutrality is moreover not
obtained at the expense of goal progress, since
$\lVert p_{\mathrm{goal}}-\mathcal{R}(\mathrm{wp}_{t-1})\rVert
=(1-\kappa)\lVert p_{\mathrm{goal}}-\mathrm{wp}_{t-1}\rVert$: the residual
distance entering $\widetilde{L}_t$ is strictly reduced, so enforcement
neither charges the metric on the step it acts nor imposes any additional
penalty on the evaluation of subsequent steps. We claim metric neutrality and
monotone reduction of the residual term only, and not a geometric
reachability guarantee, which \eqref{eq:lidar_summary} does not support. The
role of $\mathcal{R}$ is to render the hazard response bounded, deterministic
and reproducible across trials, a guarantee unavailable to the prompting-only
baselines evaluated in this work, which register the same hazard
classification but possess no pathway to act on it. The metric therefore admits its supremum for a
geometrically direct command sequence and does not by itself encode a per-step
kinematic bound. The navigation problem solved by LAPF is
\begin{equation}
\min_{\pi}\;\;\mathbb{E}\bigl[\widetilde{L}_{T_{\max}}\bigr]
\label{eq:objective}
\end{equation}
subject to the goal-reaching condition
\begin{equation}
\bigl\lVert \mathrm{wp}_{T_{\max}}-p_{\mathrm{goal}}\bigr\rVert\le\varepsilon
\label{eq:goal_condition}
\end{equation}
and to the enforcement condition
\begin{equation}
h_t\!=\!1\!\!\Longrightarrow\;\!
\!\!\lVert\mathrm{wp}_t-\mathrm{wp}_{t-1}\rVert
=\kappa\lVert p_{\mathrm{goal}}-\mathrm{wp}_{t-1}\rVert,\!\!\!\!\!\!\!\!
\qquad \forall\,t\le T_{\max},
\label{eq:enforcement_condition}
\end{equation}
which requires the hazard response to be a prescribed, model-independent
displacement rather than an arbitrary model-authored coordinate, and which
\eqref{eq:executed_wp} satisfies by construction. Condition
\eqref{eq:goal_condition} is evaluated at the terminal step of the fixed
horizon rather than used as a stopping rule, so that all methods are scored
over an identical step budget, the residual term of $\widetilde{L}_t$
penalizing any trajectory that has not met it.
\begin{table*}[!t]
\caption{Summary of the Proposed LAPF Framework Modules}
\label{tab:lapf_modules}
\centering
\footnotesize
\renewcommand{\arraystretch}{1.25}
\setlength{\tabcolsep}{5pt}
\begin{tabularx}{\textwidth}{|p{2.3cm}|p{3.3cm}|p{2.8cm}|X|}
\hline
\textbf{Module} &
\textbf{Input} &
\textbf{Output} &
\textbf{Main Functions} \\
\hline

\textbf{Perception} &
RGB image, LiDAR point cloud, and 6-DoF pose &
RGB image and symbolic state &
Registers and synchronizes RGB images for native VLM inference, converts LiDAR and localization data into a compact symbolic representation, and summarizes traversability, obstacle proximity, environmental semantics, and navigation constraints to provide situational awareness for downstream reasoning. \\
\hline

\textbf{Memory} &
Current observation, executed waypoint, and execution outcome &
Retrieved navigation experience &
Stores one navigation episode per decision step together with an LLM-generated reflection, retrieves relevant historical experiences, and maintains persistent memory to support temporal continuity, long-term spatial awareness, and adaptive reasoning. \\
\hline

\textbf{Planning} &
Current observation and retrieved memory &
Candidate waypoint &
Performs high-level multimodal reasoning using CoT to generate one candidate waypoint per decision step, together with any required corrective tool calls, according to the current scene and historical context. \\
\hline

\textbf{Action} &
Candidate waypoint and agent tool calls &
Executable waypoint &
Coordinates external tool invocation, executes discretionary waypoint-adjustment operations when required, and translates high-level planning decisions into executable navigation commands. \\
\hline

\end{tabularx}
\end{table*}
\section{Proposed LAPF} \label{sec5}

This section presents the proposed LAPF framework for autonomous UAV navigation. We first describe its overall architecture and operational workflow, followed by the agent's perception, memory, planning, and action modules. Finally, the structured prompting strategy and implementation details are introduced to illustrate how LAPF performs adaptive, closed-loop navigation.

LAPF provides a framework for autonomous UAV navigation using an LLM agent. The system comprises a UAV tasked with navigation in the considered environment, while an edge server hosts the agent and supports memory updates.

LAPF casts navigation as a closed loop rather than a single observation-to-waypoint mapping, with five stages per step $t$ tied to the formulation of Section~\ref{sec:problem}:

\begin{itemize}
\item \textbf{Perception}: produces the observation $\mathcal{O}_t=\{\mathcal{I}_t,\sigma_t,\mathcal{L}_t\}$ of \eqref{eq:obs_model}, keeping the RGB frame as an image and reducing the LiDAR return and pose to a symbolic state.
\item \textbf{Memory}: retrieves the $k=3$ most recent episodes $M_t$, each carrying an executed waypoint, its outcome, and a reflection, supplying temporal context a single pass lacks.
\item \textbf{Planning}: evaluates the policy $\pi:(\mathcal{P}_{sys},\mathcal{G},\mathcal{O}_t,M_t,\phi_t)\mapsto(\widehat{\mathrm{wp}}_t,\mathrm{Tools}_t)$ of \eqref{eq:policy}, conditioning the proposed waypoint jointly on observation, memory, and feedback.
\item \textbf{Action}: does not execute $\widehat{\mathrm{wp}}_t$ directly; the guaranteed-invocation hazard test of \eqref{eq:hazard_indicator} runs first, a registered hazard is replaced by the bounded operator $\mathcal{R}$ of \eqref{eq:replan}, and the executed waypoint follows the shielded composition of \eqref{eq:executed_wp}. This is the only stage that performs external tool invocation.
\item \textbf{Feedback}: scores the step, updates $\phi_t$, and writes the episode back into $M_t$, closing the loop.
\end{itemize}

The non-agentic baselines collapse this loop to the special case $M_t=\varnothing,\ \phi_t=\varnothing$, receiving the identical observation and the same operators, so the measured advantage isolates the memory and feedback channels rather than reasoning depth.

\par
Fig.~\ref{fig:digital567} illustrates the end-to-end operation of the proposed LAPF framework, integrating multimodal perception, memory-augmented reasoning, and tool-enforced action within a closed-loop navigation cycle. The process begins with the acquisition of multimodal environmental observations. The full sensor suite $\mathcal{O}_t^{\mathrm{full}}$ of \eqref{eq:sensor_suite} is available in UAVScenes; the pipeline evaluated in this work instantiates the subset $\mathcal{O}_t$ of \eqref{eq:obs_model}, as detailed in Table~\ref{tab:instantiation}. The depth and semantic channels are reserved for future integration and are therefore not shown in the figure. The scenario evaluated in Section~\ref{sec6} uses injected clearance violations, as described there.

\par
The pair $(\mathcal{I}_t,\,\text{symbolic prompt})$ constitutes the interface between the perception pipeline and the agent backbone, referred to as the Cognitive Navigation Core. The framework is organized into four interdependent modules, namely Perception, Memory, Planning and Action, summarized in Table~\ref{tab:lapf_modules}; Fig.~\ref{fig:digital567} arranges them by execution stage. The Perception module registers and time-matches the RGB frame for native ingestion by the VLM backbone, and separately reduces the LiDAR point cloud, obstacle range, and localization state into the compact symbolic representation that accompanies the image in the same inference call. This module provides step-synchronous situational awareness to the agent by summarizing traversability conditions, collision risks, environmental semantics, and navigation constraints, so that downstream reasoning is based on up-to-date environmental observations. The Memory module stores one episode per decision step, comprising the observation, the executed waypoint, the execution outcome, and an LLM-generated reflection on the discrepancy between the two, and maintains a persistent state representation that enables the LLM agent to recall prior navigation experiences. This memory-enhanced mechanism supports temporal continuity, adaptive reasoning, and long-term spatial awareness, allowing the agent to adapt navigation decisions over time through accumulated environmental knowledge. These two modules jointly inform the Planning module, which performs high-level CoT reasoning and autonomous decision-making to support UAV path planning and trajectory generation using techniques such as ICL and CoT reasoning. Based on the current scene understanding and historical context, the module generates a single candidate waypoint per decision step, aiming to minimize path length while supporting successful mission completion. The action module translates the planned navigation strategies into executable UAV control commands, such as waypoint updates, through textual outputs, API calls, or low-level navigation instructions. Table~\ref{tab:lapf_modules} provides a summary of the Proposed LAPF Framework Modules.

\par
Tool invocation is mediated through a dedicated external tool suite, comprising an Obstacle Checker, Trajectory Validator, Localization service, Simulator, Vector Database, and Knowledge Base, enabling the agent to evaluate candidate trajectories against safety and feasibility constraints prior to execution. The evaluation reported in Section~\ref{sec6} instantiates the two tiers of this suite that bear on the reported metrics, namely the guaranteed-invocation Obstacle Checker and the discretionary waypoint adjuster; the remaining interfaces are defined by the architecture and left for future instantiation. Upon successful validation, the resulting waypoint is transmitted to the UAV controller for execution; the waypoint sequence $\{\mathrm{wp}_t\}$ accumulates in $\mathcal{T}_{log}$ over the horizon. This constitutes a single decision cycle within the broader closed-loop autonomous navigation process depicted at the bottom of Fig.~\ref{fig:digital567}, wherein the UAV iteratively perceives its environment, encodes the non-visual channels into a structured prompt that accompanies the RGB frame, reasons over the resulting multimodal input using both memory and external tools, generates and refines a candidate trajectory, validates that trajectory for safety and feasibility, executes the resulting validated path, observes the subsequent environmental state $\mathcal{O}_{t+1}$, and updates its memory to reflect the outcome of the completed action. This iterative perception-reasoning-action cycle enables LAPF to adapt its navigation strategy in response to evolving environmental conditions without requiring gradient-based retraining of the underlying language model, in contrast to conventional reinforcement-learning-based UAV navigation approaches during execution. The cycle described here is the general operation of the architecture; under the replay-based protocol described in Section~\ref{sec6} the recorded observation stream is fixed, so the executed waypoint does not alter subsequent observations; all methods therefore receive an identical observation sequence.

Algorithm~\ref{alg:lapf} details the operational flow of the proposed LAPF framework. The procedure takes as input the UAVScenes dataset $\mathcal{D}_{\!S}$, a mission specification $\mathcal{G}$, the UAV $U$, an edge server $\mathcal{E}$ hosting the LLM agent, a system prompt $\mathcal{P}_{sys}$, and a maximum step budget $T_{\max}$. It initializes an empty memory store $\mathcal{M}$ and an empty trajectory log $\mathcal{T}_{log}$, and subsequently enters a fixed-horizon control loop that executes while $step < T_{\max}$ (lines~1--4).

\textbf{\textit{Perception}} (lines~5--8): at each decision step, the algorithm retrieves the current frame's multimodal data directly from the dataset, comprising the RGB image $\mathcal{I}_t$, the LiDAR point cloud $\mathcal{P}_t$, and the 6-DoF pose $\mathcal{L}_t$. The point cloud is reduced to the summary statistic $\sigma_t$ of \eqref{eq:lidar_summary}, and the resulting quantities are assembled into the observation $\mathcal{O}_t$ of \eqref{eq:obs_model}. The non-visual channels $(\sigma_t,\mathcal{L}_t)$ are then rendered into the symbolic state $s_t$, whereas $\mathcal{I}_t$ is retained as an image and is not textualized.

\textbf{\textit{Memory}} (lines~9--11): the memory store $\mathcal{M}$ is queried for the $k$ most recent prior episodes $M_t$ (with $k=3$), so that retrieval is recency-based rather than similarity-based, enabling the agent to condition its subsequent reasoning on previously encountered trajectories, failures, and successful outcomes. The scalar performance-feedback signal $\phi_t$ of \eqref{eq:policy} is computed over the trailing window of path-length statistics.

\textbf{\textit{Planning}} (lines~12--14): the mission goal $\mathcal{G}$, the symbolic state $s_t$, the retrieved memory $M_t$, and the feedback signal $\phi_t$ are combined into a context $Ctx$. This context, together with the system prompt $\mathcal{P}_{sys}$ and the image $\mathcal{I}_t$, is passed to the agent backbone $\mathcal{E}.\mathrm{Agent}(\cdot)$, which returns a reasoning trace $Thought$, a proposed waypoint $Waypoint$, and a set of tool calls $ToolCalls$ in a single multimodal inference call.

\textbf{\textit{Action}} (lines~15--24): Hazard evaluation is performed with guaranteed invocation at every decision step, so that its activation does not depend on whether the corresponding tool is named in the tool set returned at line~14. It evaluates the sensed clearance $c_t$ against the threshold
$d_{\mathrm{safe}}$, returning the hazard indicator $h$ of
\eqref{eq:hazard_indicator}. If $h=1$, the deterministic corrective operator
\eqref{eq:replan} overwrites $Waypoint$ with a value computed from the
commanded state and the mission goal; otherwise the stability operator
\eqref{eq:clamp} is applied to the model-authored proposal. If \texttt{WaypointAdjust} appears in $ToolCalls$, the hold-position operator $\mathcal{H}$ of \eqref{eq:hold} is subsequently applied, completing the ordered composition of \eqref{eq:executed_wp}.

\textbf{\textit{Execution}} (lines~25--26): the resulting waypoint, potentially modified by the Action step, is transmitted to the UAV via $U.\mathrm{execute}(Waypoint)$, and outcome feedback is subsequently retrieved via $U.\mathrm{get\_feedback}()$.

\textbf{\textit{Memory Update}} (lines~27--30): the edge server evaluates the discrepancy between the original reasoning trace $Thought$ and the observed $Outcome$, producing a $Reflection$. The complete episode $(\mathcal{O}_t, Waypoint, Outcome, Reflection)$ is then stored in memory $\mathcal{M}$ for retrieval in subsequent iterations, and the UAV's current pose is appended to the trajectory log $\mathcal{T}_{log}$.

The step counter is incremented (line~31) and the loop repeats until $T_{\max}$ steps have elapsed, at which point the complete trajectory log $\mathcal{T}_{log}$ is returned.

\begin{algorithm}[t]
\scriptsize
\caption{Agentic LAPF Framework}
\label{alg:lapf}

\begin{algorithmic}[1]

\Require UAVScene dataset $\mathcal{D}_{\!S}$, mission $\mathcal{G}$, UAV $U$, edge server $\mathcal{E}$, system prompt $\mathcal{P}_{sys}$, maximum steps $T_{\max}$
\Ensure Navigation trajectory $\mathcal{T}_{log}$

\State Initialize memory $\mathcal{M}\gets\emptyset$
\State Initialize trajectory log $\mathcal{T}_{log}\gets\emptyset$
\State $step\gets0$

\While{$step<T_{\max}$}

\State \textbf{Perception}

\State $(\mathcal{I}_t,\mathcal{P}_t,\mathcal{L}_t)\gets\mathcal{D}_{\!S}[step]$
\State $\sigma_t\gets\mathrm{summarize}(\mathcal{P}_t)$;\quad
       $\mathcal{O}_t\gets\{\mathcal{I}_t,\sigma_t,\mathcal{L}_t\}$
\State $s_t\gets \mathrm{encode\_to\_text}(\sigma_t,\mathcal{L}_t)$
\Comment{$\mathcal{I}_t$ kept as image}

\State \textbf{Memory}

\State $M_t\gets\mathcal{M}.\mathrm{retrieve\_recent}(k=3)$
\State $\phi_t\gets\mathrm{feedback}(\mathcal{T}_{log})$
\Comment{trailing window}

\State \textbf{Planning}

\State $Ctx\gets\{\mathcal{G},s_t,M_t,\phi_t\}$

\State $(Thought,Waypoint,ToolCalls)\gets
\mathcal{E}.\mathrm{Agent}(\mathcal{P}_{sys},\mathcal{I}_t,Ctx)$

\State \textbf{Action}

\State $h\gets \mathbb{1}[\,c_t\le d_{\mathrm{safe}}\,]$
\Comment{guaranteed invocation}

\If{$h=1$}
    \State $Waypoint\gets \mathcal{R}(\mathrm{wp}_{t-1})$
    \Comment{Eq.~\eqref{eq:replan}}
\Else
    \State $Waypoint\gets \mathcal{C}(Waypoint)$
    \Comment{Eq.~\eqref{eq:clamp}}
\EndIf

\If{\texttt{WaypointAdjust}$\,\in ToolCalls$}
    \State $Waypoint\gets \mathcal{H}(Waypoint)$
    \Comment{Eq.~\eqref{eq:hold}}
\EndIf

\State $U.\mathrm{execute}(Waypoint)$
\State $Outcome\gets U.\mathrm{get\_feedback}()$

\State \textbf{Memory Update}

\State $Reflection\gets
\mathcal{E}.\mathrm{Reflect}(Thought,Outcome)$

\State $\mathcal{M}.\mathrm{store}
(\mathcal{O}_t,Waypoint,Outcome,Reflection)$

\State $\mathcal{T}_{log}.\mathrm{append}(U.\mathrm{pose}())$

\State $step\gets step+1$

\EndWhile

\Return $\mathcal{T}_{log}$

\end{algorithmic}
\end{algorithm}

\begin{figure*}[t]
\centering
\begin{tcolorbox}[
enhanced,
width=\textwidth,
colback=gray!3,
colframe=black!50,
boxrule=0.4pt,
arc=0mm,
left=1mm,
right=1mm,
top=0.8mm,
bottom=0.8mm,
title={LAPF Prompt Template},
fonttitle=\bfseries\footnotesize,
colbacktitle=gray!15,
coltitle=black
]

\scriptsize
\renewcommand{\arraystretch}{1.05}

\begin{tabularx}{\textwidth}{@{}p{0.19\textwidth}X@{}}

\textbf{System Prompt} &
\ttfamily
You are an expert LAPF UAV navigation agent. Use CoT reasoning for safe and efficient path planning while minimizing path length.
\\
\hline

\textbf{Mission} &
\ttfamily
Mission=\{mission\}\quad
Goal=(gx,gy,gz)\quad
Current=(x,y,z)
\\
\hline

\textbf{Image input} &
\ttfamily
$\langle$image$\rangle$ $\mathcal{I}_t$ (\vlmres{}) --- passed natively to the
VLM backbone as a separate content block, not as text
\\
\hline

\textbf{Symbolic State} &
\ttfamily
Pose=(x,y,z),
RTK=(lat,lon,alt),
LiDAR=\{n\_points\}pts,
Clearance=\{c\}m,
range=[rmin,rmax],
lateral=[bmin,bmax]
\\
\hline

\textbf{Retrieved Memory} &
\ttfamily
Waypoint=\{wp\};
Outcome=\{outcome\};
Reflection=\{reflection\};
$k$ most recent episodes ($k=3$)
\\
\hline

\textbf{Performance} &
\ttfamily
Error=X m;
Trend=degrading;
Efficiency=Y\%;
Large deviation $\rightarrow$ choose a more direct waypoint.
\\
\hline

\textbf{Reasoning} &
1) Assess current state;
2) Retrieve memory;
3) Plan safe \& efficient waypoint;
4) Select required tools.
\\
\hline

\textbf{JSON Output} &
\ttfamily
\{"situation\_analysis":"...", "thought":"...", "waypoint":[x,y,z], "tool\_calls":["Tool1","Tool2"]\}
\\

\end{tabularx}
\end{tcolorbox}

\caption{Multimodal prompt supplied to the VLM backbone at each planning cycle. The RGB frame is passed natively as an image block; all remaining fields are text, instantiated at runtime from $\sigma_t$, $\mathcal{L}_t$, the mission goal, the retrieved episodic memory $M_t$, and the feedback signal $\phi_t$. The returned JSON is parsed into \emph{Thought}, \emph{Waypoint}, and \emph{ToolCalls} by the Planning module.}

\label{fig:prompt_template}

\end{figure*}

To ground the agent's reasoning in a machine-parseable format at every decision step, LAPF employs a fixed structured template rather than free-form querying. Fig.~\ref{fig:prompt_template} illustrates this template, comprising seven input fields and a JSON output format. A system prompt fixes the agent's role and CoT objective. A mission field gives the goal coordinates and current position. The RGB frame is supplied as a separate image block. A symbolic state field carries the pose, the RTK geolocation, and the summarized LiDAR return, namely the point count, the clearance $c_t$, and the range and lateral bounds. A retrieved-memory field lists the $k$ most recent episodes. A performance field reports the current path-length error, trend, and efficiency score. A reasoning field instructs the agent to assess the situation, consult memory, plan a candidate waypoint, and select any required tools.

To allow the Planning module to parse \emph{Thought}, \emph{Waypoint} and
\emph{ToolCalls} deterministically at every step, the template constrains the
output to a single raw JSON object with no surrounding text. The reasoning
trace is thereby retained as the \texttt{thought} field of the structured
response rather than as unconstrained free text, so that CoT
content remains available for the reflection step without compromising
machine-parseability.

\begin{table}[t]
\caption{LAPF implementation hyperparameters}
\label{tab:hyperparams}
\centering
\footnotesize
\setlength{\tabcolsep}{4pt}
\renewcommand{\arraystretch}{1.1}

\begin{tabularx}{\columnwidth}{@{}p{0.42\columnwidth}X@{}}
\hline
\textbf{Parameter} & \textbf{Value} \\ \hline

\multicolumn{2}{@{}l}{\textit{LLM inference}} \\

Backbone & \vlm{} (open-weight VLM) \\
Serving backend & \vlmbackend{} \\
Image input resolution & \vlmres{} (resampled from $2448\times2048$) \\
Temperature & 0.7 \\
Max tokens (planning) & 8192 \\
Max tokens (reflection) & 150 \\
Call timeout (planning) &
600 s (LAPF), 120 s (baselines); scaled to context length \\
\hline
\multicolumn{2}{@{}l}{\textit{Mission configuration}} \\

Decision horizon $T_{\max}$ & 110 steps \\
Dataset frame skip & 20 \\
Frames in evaluated subset & \nframes{} \\
Independent trials per configuration & 3 \\

\hline
\multicolumn{2}{@{}l}{\textit{Memory \& feedback}} \\

Retrieved past experiences $k$ & 3 \\
Feedback trend window & 5 steps \\
Critical-deviation threshold & $>10.0$ m \\
Warning-deviation threshold & $>5.0$ m \\
High-detour-ratio threshold & $>1.5\times$ \\
Low-efficiency threshold & $<60\%$ \\
High-efficiency threshold & $>85\%$ \\

\hline
\multicolumn{2}{@{}l}{\textit{Safety \& action}} \\

Clearance threshold $d_{\mathrm{safe}}$ & $5.0$ m \\
Near-goal feedback trigger & $<2.0$ m \\
Near-goal step bound $\delta_{\max}$ & $2.0$ m \\
Replan contraction $\kappa$ & $0.5$ \\

\hline
\multicolumn{2}{@{}l}{\textit{Reliability}} \\

Max retries per LLM call & 20 \\
Retry backoff & $\min(2^{\text{attempt}},15)$ s \\

\hline
\end{tabularx}
\end{table}

\section{Numerical Results and Discussions} \label{sec6}
This section evaluates the performance of the proposed LAPF framework through numerical experiments on the UAVScenes dataset. We first describe the implementation settings and baseline methods, and then present comparative results under different navigation scenarios to assess path efficiency, safety, and overall navigation performance.
\subsection{Implementation of LAPF}
LAPF is implemented in Python and evaluated on a MacBook Air equipped with an Apple M5 processor and 16 GB unified memory under macOS. The framework uses Qwen2-VL-7B-Instruct as the open-weight vision-language backbone, served locally through 'mlx-vlm', enabling the complete evaluation to run on a single laptop without relying on proprietary large-scale models. RGB inputs are resized from their original 2448×2048 resolution to 1024×768 before inference to control the visual token budget. The structured prompt template enforces JSON-based outputs, with invalid responses handled through retries. A geometric fallback mechanism is monitored through the \texttt{llm\_fallback\_steps} log; across the two scenarios and three trials, LAPF executes $2\times3\times110=660$ decision steps in total, over which this fallback was never activated. The evaluation uses three independent trials per configuration with a maximum horizon of 110 steps on the AMtown03 subset containing 41 frames. Since the decision horizon $T_{\max}=110$ exceeds the $41$-frame observation subset, the recorded stream is cycled once exhausted, so that every method is scored over an identical $110$-step budget irrespective of sequence length. The remaining implementation parameters, including inference settings, memory configuration, feedback thresholds, and safety-related parameters, are summarized in Table~\ref{tab:hyperparams}. All methods follow the same observation schedule, ensuring a fair comparison where performance differences arise from decision-making strategies rather than input variations. 

\subsection{Baselines Description}

\textbf{Pure LLM} represents a non-agentic baseline that directly generates navigation decisions from the input state without external feedback, memory, or iterative action correction.
\par

\textbf{CoT LLM} represents a reasoning-enhanced but still non-agentic baseline, in which the model is prompted to perform step-by-step reasoning before generating navigation actions. 
\par
All three methods are driven by the same \vlmshort{} backbone under identical decoding settings. Crucially, all three receive the \emph{identical} multimodal observation: the RGB frame $\mathcal{I}_t$ is passed natively as an image input in every configuration, and the symbolic fields derived from $\sigma_t$ and $\mathcal{L}_t$ are byte-identical across the three prompts, as is the output schema.  All three are additionally evaluated under the same near-goal stability operator $\mathcal{C}$ of \eqref{eq:clamp} and the same hold-position operator $\mathcal{H}$ of \eqref{eq:hold}, which are applied uniformly rather than only to the proposed method.

\subsection{Open-Field Navigation (AMtown03)}

\begin{figure*}[t]
\centering
\includegraphics[width=18cm]{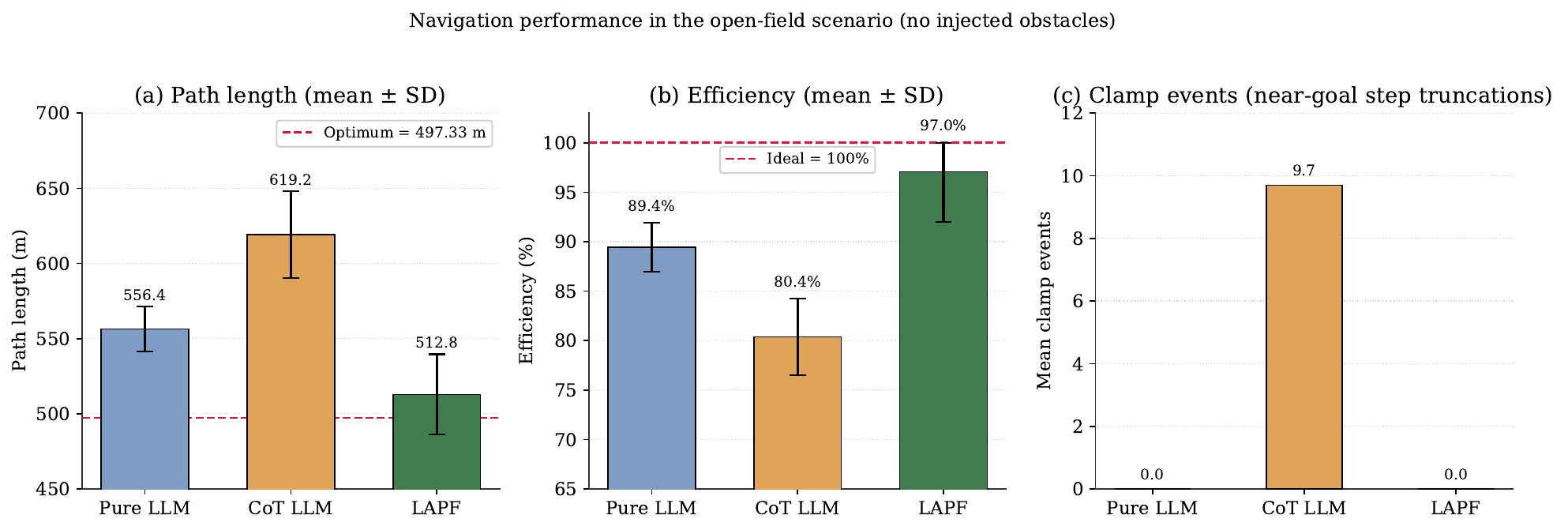}
\caption{Open-field scenario (AMtown03, no injected obstacles). Panels (a) and (b) show path length and efficiency (mean $\pm$ SD) across three trials per method, relative to the $497.33$\,m straight-line optimum shown as a dashed red line. Panel (c) shows mean clamp events (near-goal step truncations). LAPF converges to near-optimal path length and efficiency with zero clamp events, while CoT LLM shows the largest deviation from optimal and the only nonzero clamp-event rate even in the absence of obstacles.}
\label{fig:open_field_summary}
\end{figure*}

\begin{figure*}[t]
\centering
\includegraphics[width=\textwidth]{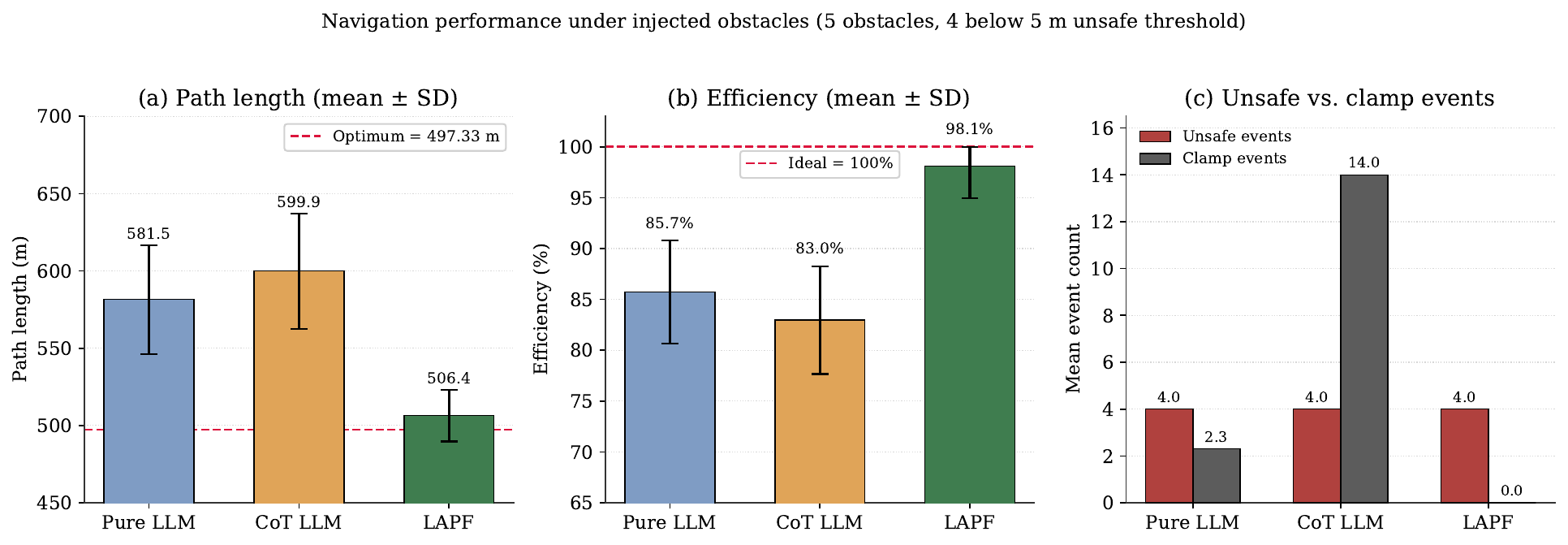}
\caption{Obstacle-injected scenario (AMtown03, five injected clearance violations, four below $d_{\mathrm{safe}}=5$\,m). Panels (a) and (b) show path length and efficiency (mean $\pm$ SD) across three trials per method, relative to the $497.33$\,m straight-line optimum shown as a dashed red line. Panel (c) shows mean unsafe-event and clamp-event counts. LAPF converges to near-optimal path length and efficiency while incurring zero clamp events, whereas CoT LLM shows the largest deviation and the highest clamp-event frequency, consistent with less regulated step proposals under obstacle constraints.}
\label{fig:obstacle_injected_summary}
\end{figure*}

Fig.~\ref{fig:open_field_summary} summarizes path length (panel a), efficiency (panel b), and clamp events (panel c) relative to the theoretical straight-line optimum (dashed red line).
\par
This scenario contains no injected obstacles, so $h_t=0$ at every step and $N_h=0$ in all trials: by \eqref{eq:executed_wp} every executed waypoint here is the model-authored proposal under the same operator $\mathcal{C}$ applied to the baselines. As shown in Fig.~\ref{fig:open_field_summary}(a), LAPF achieves the shortest mean path length (512.83 $\pm$ 26.78$\,$m), closely tracking the optimal-path reference line, with T1 ($497.33\,$m) and T2 ($497.40\,$m) nearly coincident with it. Since $N_h=0$, the coincidence of T1 with $L^{\star}$ reflects a collinear command sequence issued by the policy itself, not a geometric fallback. Pure LLM yields a longer mean path length of 556.37 $\pm$ 15.02$\,$m, while CoT LLM exhibits the largest deviation from the optimal trajectory, averaging 619.16 $\pm$ 29.01$\,$m. LAPF operates through a closed-loop agent architecture that enables the LLM to perceive outcomes, maintain memory of previous actions, and iteratively refine future decisions. The reduced path length therefore reflects not only improved reasoning, but also the ability of the LLM agent to ground its decisions in feedback and continuously correct its behavior.

This trend is further supported by Fig.~\ref{fig:open_field_summary}(b), where LAPF achieves the highest mean efficiency (97.05 $\pm$ 5.09\%) and the lowest mean detour ratio (1.032 $\pm$ 0.056), approaching the 100\% efficiency reference. In comparison, Pure LLM achieves intermediate performance (89.43 $\pm$ 2.45\% efficiency, with a 1.119 detour ratio), while CoT LLM performs worst (80.37 $\pm$ 3.85\% efficiency, 1.246 detour ratio). 
The observed advantage is consistent with interaction between reasoning and externalized control mechanisms, where observations, memory, and corrective actions allow the system to adapt its behavior rather than relying on prompting alone.

Fig.~\ref{fig:open_field_summary}(c) provides additional evidence of this distinction. Both Pure LLM and LAPF achieve zero clamp events across all trials, indicating stable waypoint generation and appropriate step-size regulation. In contrast, CoT LLM produces an average of 9.7 clamp events despite operating in an obstacle-free environment. Without an execution-feedback loop, the model may generate actions that are internally plausible but poorly calibrated. LAPF avoids this issue by coupling LLM decision-making with explicit action constraints and feedback, allowing the agent to better regulate waypoint execution.
\par

The near-optimal values in panels (a) and (b) warrant explicit interpretation. By \eqref{eq:path_length}, $\eta_t$ attains unity exactly when the commanded sequence is collinear with, and monotone along, the start--goal segment, so a value of $100\%$ is admissible by construction and is not an artefact of the measurement. It does, however, indicate that the metric rewards geometric directness without separately certifying per-step kinematic realizability, since \eqref{eq:executed_wp} constrains step magnitude only in the terminal phase through $\mathcal{C}$. Path efficiency is therefore reported here as a measure of trajectory directness relative to the straight-line optimum, and is complemented by the clamp count $N_c$ of \eqref{eq:clamp_count}, which is sensitive precisely to unregulated step magnitudes. Introducing a global displacement bound $\lVert \mathrm{wp}_t-\mathrm{wp}_{t-1}\rVert\le\Delta_{\max}$ applied uniformly to all methods, so that directness and kinematic feasibility are scored jointly, is identified as future work.
\par

The ability to maintain memory, observe execution outcomes, enforce constraints, and revise actions enables LAPF to produce more efficient and stable trajectories. These results suggest that effective LLM-assisted autonomy depends not only on model reasoning but also on the surrounding agent architecture that converts reasoning into adaptive behavior.

\subsection{Obstacle-Injected Navigation (AMtown03)}

Fig.~\ref{fig:obstacle_injected_summary} extends the open-field evaluation by injecting five obstacles into the environment, and compares Pure LLM, CoT LLM, and LAPF across three trials (T1--T3) relative to the same 497.33\,m straight-line optimum. Obstacles are injected by overriding the sensed clearance $c_t$ at five pre-selected decision steps, identical across methods and trials. Four of these are set below $d_{\mathrm{safe}}=5$\,m and therefore register as hazards. Injection is applied to the observation stream rather than to the commanded state, which preserves the common-observation control and accounts for the identical unsafe-event count in panel~(c). Controlled injection is also what exercises the enforcement pathway at all, since the recorded sequence is flown as a nadir survey and $c_t$ remains above $15$\,m in every frame, so it never approaches $d_{\mathrm{safe}}=5$\,m at any point of the flight. The results below therefore establish mechanism correctness under a controlled stimulus. Evaluation against naturally occurring hazards requires a lower-clearance sequence or a forward-facing sensor and is identified as future work.

\par
Fig.~\ref{fig:obstacle_injected_summary} further demonstrates the importance of agentic capabilities when an LLM-assisted navigation system is exposed to increased environmental complexity. The obstacle-injected scenario introduces a setting where a single-pass generation strategy is insufficient: the system must not only produce a trajectory but also incorporate environmental constraints, regulate actions, and adapt its decisions based on feedback.
\par
As shown in Fig.~\ref{fig:obstacle_injected_summary}(a), LAPF achieves the shortest mean path length (506.37 $\pm$ 16.68$\,$m), remaining closest to the optimal-path reference line despite the presence of additional obstacles. In comparison, Pure LLM achieves 581.51 $\pm$ 35.20$\,$m, while CoT LLM reaches 599.89 $\pm$ 37.20$\,$m, showing the largest deviation from the optimum. Although all methods experience some increase in path length compared with the open-field condition due to obstacle avoidance requirements, LAPF exhibits almost no degradation. This robustness is consistent with the agentic structure enabling the LLM to integrate observations, maintain state information, and revise its trajectory when environmental constraints change, rather than relying on a fixed pre-generated plan.

\par
The same trend appears in Fig.~\ref{fig:obstacle_injected_summary}(b), where LAPF maintains an efficiency of 98.09 $\pm$ 3.17\% and a detour ratio of 1.020 $\pm$ 0.034, closely matching its obstacle-free performance. In contrast, Pure LLM and CoT LLM undergo larger efficiency reductions, achieving 85.73 $\pm$ 5.05\% and 82.96 $\pm$ 5.30\%, respectively. These results suggest that additional reasoning depth alone is insufficient for effective obstacle handling, which also requires interaction and adaptation. Although CoT LLM performs additional internal reasoning, it lacks the external feedback mechanism needed to assess whether its generated actions remain appropriate under changing conditions. LAPF’s closed-loop agent architecture grounds reasoning in perception, memory, and action correction, enabling it to preserve high path efficiency even as the environment becomes more challenging. The stability of $\eta_t$ under obstacle injection is a designed property rather than an incidental one. By Property~1 the enforcement pathway is metric-neutral by construction, so the efficiency reported here measures the agentic decision policy and not the shield, and enforcement is instead assessed through $N_h$ and the clamp count. This separation is what makes the present comparison interpretable. By Property~1 the hazard-classified steps contribute exactly zero to $\eta_t$, so the $93.5$\,m and $75.1$\,m advantages of LAPF over CoT LLM and Pure LLM are predominantly attributable to the decision policy, which under an identical backbone, identical decoding settings and an identical observation stream is most consistently explained by the agentic mechanism that distinguishes LAPF, namely the joint conditioning of every proposal on episodic memory and on the performance-feedback signal $\phi_t$.

\par
Fig.~\ref{fig:obstacle_injected_summary}(c) shows identical unsafe-event counts across all methods (mean 4.0), corresponding to the same set of injected obstacles below the safety threshold. This identity is intentional and serves as an experimental control: it certifies that the three methods were exposed to the same hazard set, so the divergence observed in the response metrics cannot be ascribed to unequal environmental difficulty. What distinguishes the methods is not how often a hazard is registered, because this is common to all methods, but whether registration is coupled to a prescribed response: LAPF converts all four instances into enforced replans through \eqref{eq:enforcement_condition}, whereas the prompting-only baselines classify the same four instances with no pathway to act on them. A more meaningful distinction appears in the clamp-event results, where LAPF maintains zero clamp events, Pure LLM shows a small increase (2.3 events), and CoT LLM increases substantially to 14.0 events. This distinction is not an artefact of the exemption of hazard steps from $\mathcal{C}$: the exemption spans at most four steps, whereas the separation from CoT LLM is $14.0$, and in the open-field scenario $N_h=0$, so $\mathcal{C}$ is active on all $110$ steps of every method and $N_c=0$ is attained by LAPF with no exemption whatsoever. The high clamp rate of CoT LLM suggests that additional reasoning does not guarantee improved action regulation. Without an execution-feedback loop, the model may produce logically plausible but poorly calibrated waypoint predictions, particularly when jointly considering goal progression and obstacle constraints.

\par
Overall, panels (a)–(c) show that LAPF’s advantage is consistent with its agentic capabilities: perceiving the environment, preserving memory, enforcing action constraints, and iteratively refining decisions. Unlike non-agentic LLM baselines, whose performance becomes less stable as task complexity increases, LAPF maintains efficient, controlled navigation under environmental perturbations. These results suggest that reliable LLM-assisted autonomy depends not only on the model’s reasoning ability but also on the surrounding agent framework that translates reasoning into adaptive, feedback-driven action.

\subsection{Discussion}
The results are best interpreted through the closed loop of \eqref{eq:policy}, in which each waypoint is conditioned not on the current observation alone but jointly on the episodic memory $M_t$ and the feedback signal $\phi_t$. The non-agentic baselines are the special case $M_t=\varnothing,\ \phi_t=\varnothing$; since all three receive the byte-identical observation and the same operators $\mathcal{C}$ and $\mathcal{H}$, the measured gap isolates these two channels rather than reasoning depth or perception quality.
\par
\emph{Role of memory.} Recency-based retrieval over the $k=3$ most recent episodes, each carrying an executed waypoint, its outcome, and a reflection on their discrepancy, lets the agent condition every proposal on how its own recent commands actually resolved. This is what turns a per-step reaction into a corrected trajectory: in the open-field case it yields near-collinear command sequences (T1 coincident with $L^{\star}$ at $N_h=0$), and under obstacle injection it preserves efficiency at $98.09\%$ rather than degrading, because past hazard episodes bias subsequent waypoints toward directions that anticipate the shield instead of repeatedly triggering it. Without $M_t$, the baselines regenerate each waypoint from scratch, a mechanism most consistent with their larger detour ratios and, for CoT LLM, the increase in clamp events from $9.7$ to $14.0$.
\par
\emph{Role of tools.} Tool use contributes a distinct and complementary effect: it separates what the agent \emph{decides} from what is \emph{guaranteed}. The guaranteed-invocation Obstacle Checker of \eqref{eq:hazard_indicator} tests clearance at every step independently of $\mathrm{Tools}_t$, so a registered hazard is always coupled to the bounded response of \eqref{eq:replan}, while the discretionary waypoint adjuster of \eqref{eq:hold} regulates step magnitude near the goal. Fig.~\ref{fig:obstacle_injected_summary}(c) shows that all methods register the same four unsafe events, yet only LAPF converts each into an enforced replan; by Property~1 this pathway is metric-neutral, so the two effects, safer response and shorter paths, are scored through separate estimators ($N_h$ versus $\eta_t$) and never conflated.
\par
Taken together, memory and feedback shorten and stabilize the trajectory, while tool-enforced action guarantees the hazard response; neither alone reproduces the full behavior. This is why additional reasoning depth, as in CoT LLM, is insufficient: without a feedback channel to test whether generated actions remain appropriate, and without an enforced path to act on detected hazards, deeper reasoning yields internally plausible but poorly calibrated waypoints, particularly when goal progression and obstacle constraints must be considered jointly. In the obstacle-injected scenario, LAPF preserves both path efficiency and trajectory quality with minimal degradation: its mean path length changes only from $512.83$\,m to $506.37$\,m and its efficiency rises from $97.05\%$ to $98.09\%$, with zero clamp events in both, whereas CoT LLM rises from $9.7$ to $14.0$ clamp events and Pure LLM from $0.0$ to $2.3$. The advantage is therefore not limited to unconstrained navigation but is sustained as complexity increases.
\par
LAPF is not intended to replace classical geometric planners when accurate maps and deterministic objectives are available; it targets environments that require semantic interpretation, adaptive replanning, and integration of heterogeneous information sources, providing a higher-level cognitive layer that formulates, revises, and validates navigation strategies. Several limitations remain: the evaluation is confined to UAVScenes and excludes dense urban settings with narrow corridors, tall buildings, and pedestrian interactions; the semantic and depth channels are not yet fused into the reasoning pipeline; and the safety layer operates on a conservative scalar clearance rather than a full 3D collision model.
Finally, the contribution of the visual channel is not isolated here: the image is a control variable rather than a treatment, since all three methods receive it identically, and an image-ablated configuration is left to future work.

\section{Conclusion} \label{sec7}

This paper presents LAPF, an LLM-agent-based framework for autonomous UAV navigation that integrates multimodal perception, episodic memory, planning, and tool-enforced action within a closed-loop architecture. Using observations from UAVScenes, LAPF combines native RGB input with symbolic LiDAR and 6-DoF pose representations through a compact open-weight VLM, enabling edge-scale multimodal navigation on a single consumer laptop. The proposed memory and action mechanisms support adaptive waypoint generation and guarantee safety-critical hazard responses during navigation.

Experimental results show that LAPF consistently outperforms conventional prompting and CoT prompting in both open-field and obstacle-injected scenarios, achieving higher path efficiency, shorter trajectories, and improved navigation stability while maintaining zero clamp events under obstacle injection. In our experiments, the most decisive advantage…of the agentic design is the coupling of memory-grounded feedback with tool-enforced action: this pairing, rather than reasoning depth alone, allows LAPF to both shorten trajectories and guarantee a bounded hazard response, and most reliably distinguishes it from prompting-only methods as task complexity grows. Although the current evaluation is limited to a single scene, one backbone, and three trials per configuration, the results demonstrate the effectiveness of integrating memory, reasoning, and tool interaction for autonomous UAV navigation. Future work will extend the framework to larger-scale evaluations, interactive closed-loop simulation, and real-world UAV deployment.
\bibliographystyle{IEEEtran}
\bibliography{references}
\end{document}